\documentclass[11pt]{article}

\usepackage{fullpage}
\usepackage[ruled, linesnumbered, vlined, commentsnumbered]{algorithm2e}
\usepackage{graphicx,subfig} 
\usepackage{amsfonts,amssymb,amsmath,amsthm,amsopn}	
\usepackage{hhline,booktabs,colortbl,multirow,tabularx,threeparttable} 
\usepackage{gensymb}

\usepackage{enumerate}
\usepackage{authblk}
\usepackage{footnote}
\usepackage{hyperref}
\usepackage{prettyref}
\usepackage{cite}
\usepackage{setspace}
\usepackage{color}
\usepackage{xcolor}  
\usepackage{scrpage2}
\usepackage{geometry}

\usepackage{tikz}
\usepackage{pgfplots}
\usetikzlibrary{positioning,shapes,shadows,arrows,calc}
\tikzstyle{component}=[rectangle, draw=black, rounded corners, fill=blue!40, drop shadow, text centered, anchor=north, text=white, minimum height=1cm]
\tikzstyle{arrow}=[->, thick]

\pgfplotsset{compat=1.12}
\usetikzlibrary{intersections}
\usetikzlibrary{pgfplots.statistics}
\usepgfplotslibrary{fillbetween}

\definecolor{myblue}{RGB}{34,31,217}
\definecolor{mycyan}{gray}{.7}
\definecolor{Gray}{gray}{0.9}

\newcommand{\pref}{\prettyref}

\newrefformat{fig}{Fig.~\ref{#1}}
\newrefformat{tab}{Table~\ref{#1}}
\newrefformat{sec}{Section~\ref{#1}}
\newrefformat{app}{Appendix~\ref{#1}}
\newrefformat{alg}{Algorithm~\ref{#1}}
\newrefformat{property}{Property~\ref{#1}}
\newrefformat{theorem}{Theorem~\ref{#1}}
\newrefformat{corollary}{Corollary~\ref{#1}}
\newrefformat{proposition}{Proposition~\ref{#1}}
\newrefformat{def}{Definition~\ref{#1}}
\newrefformat{eq}{equation~(\ref{#1})}

\usepackage{lscape}

\begin{document}

\title{\vspace{-1ex}\LARGE\textbf{Visualisation of Pareto Front Approximation: A Short Survey and Empirical Comparisons\footnote{This manuscript is submitted for possible publication. The reviewer can use this manuscript in peer review.}}}

\author[1]{\normalsize Huiru Gao}
\author[1]{\normalsize Haifeng Nie}
\author[2]{\normalsize Ke Li}
\affil[1]{\normalsize School of Computer Science and Engineering\\ University of Electronic Science and Technology of China}
\affil[2]{\normalsize Department of Computer Science, University of Exeter}
\affil[$\ast$]{\normalsize Email: \texttt{k.li@exeter.ac.uk, \{huir.gao, seapeak.nie\}@gmail.com}}

\date{}
\maketitle

\vspace{-3ex}
{\normalsize\textbf{Abstract: } }Visualisation is an effective way to facilitate the analysis and understanding of multivariate data. In the context of multi-objective optimisation, comparing to quantitative performance metrics, visualisation is, in principle, able to provide a decision maker better insights about Pareto front approximation sets (e.g. the distribution of solutions, the geometric characteristics of Pareto front approximation) thus to facilitate the decision-making (e.g. the exploration of trade-off relationship, the knee region or region of interest). In this paper, we overview some currently prevalent visualisation techniques according to the way how data is represented. To have a better understanding of the pros and cons of different visualisation techniques, we empirically compare six representative visualisation techniques for the exploratory analysis of different Pareto front approximation sets obtained by four state-of-the-art evolutionary multi-objective optimisation algorithms on the classic DTLZ benchmark test problems. From the empirical results, we find that visual comparisons also follow the \textit{No-Free-Lunch} theorem where no single visualisation technique is able to provide a comprehensive understanding of the characteristics of a Pareto front approximation set. In other words, a specific type of visualisation technique is only good at exploring a particular aspect of the data.



\section{Introduction}
\label{sec:introduction}

Multi-objective optimisation problems (MOPs), which consider simultaneously optimising more than one conflicting objective function, are ubiquitous across the breadth of science~\cite{Coello00}, engineer~\cite{MontanoCM12,LiCFY18} and economics~\cite{PonsichJC13}. Specifically, this paper considers the MOP formally defined as:
\begin{equation}
\begin{array}{l l}
\mathrm{minimize} \quad \mathbf{F}(\mathbf{x})=(f_{1}(\mathbf{x}),\cdots,f_{m}(\mathbf{x}))^{T}\\
\mathrm{subject\ to} \quad \mathbf{x} \in\Omega
\end{array}
\label{MOP}
\end{equation} 
where $\mathbf{x}=(x_1,\ldots,x_n)^T$ is a candidate solution, and $\Omega=[x_i^L,x_i^U]^n\subseteq\mathbb{R}^n$ defines the search (or decision variable) space. $\mathbf{F}:\Omega\rightarrow\mathbb{R}^m$ constitutes $m$ conflicting objective functions, and $\mathbb{R}^m$ is the objective space. Given two solutions $\mathbf{x}^1$, $\mathbf{x}^2\in\Omega$, we said that $\mathbf{x}^1$ dominates $\mathbf{x}^2$ (denoted as $\mathbf{x}\preceq\mathbf{x}^2$) in case $\mathbf{F}(\mathbf{x}^1)$ is not worse than $\mathbf{F}(\mathbf{x}^2)$ in any individual objective and it at least has one better objective. A solution $\mathbf{x}^{\ast}$ is Pareto-optimal with respect to (\ref{MOP}) in case $\nexists\mathbf{x}\in\Omega$ such that $\mathbf{x}\preceq\mathbf{x}^{\ast}$. The set of all Pareto-optimal solutions is called the Pareto set (PS). Accordingly, $PF=\{\mathbf{F}(\mathbf{x})|\mathbf{x}\in PS\}$ is called the Pareto front (PF).

Evolutionary algorithms (EAs), which work with a population of solutions, are widely accepted as a major approach for multi-objective optimisation. In the past three decades and beyond, significant efforts have been devoted on the developments of evolutionary multi-objective optimisation (EMO) algorithms, such as fast non-dominated sorting genetic algorithm (NSGA-III)~\cite{DebAPM02}, indicator-based EA (IBEA)~\cite{ZitzlerK04} and multi-objective EA based on decomposition (MOEA/D)~\cite{ZhangL07}, to find a set of well-converged and well-diversified efficient solutions that approximate the whole PF. Nevertheless, the ultimate goal of MO is to help the DM find solution(s) that meet her/his preferences. Providing the DM with a large amount of raw data inarguably incurs severe cognitive obstacle for decision-making, especially when having many objectives (i.e. $m>3$). 
 Although there are some performance metrics that can evaluate the quality of a PF approximation set (e.g. inverted generational distance (IGD)~\cite{BosmanT03} and Hypervolume (HV)~\cite{ZitzlerT99}), it is still not intuitive to help the DM understand the trade-off relationship among objectives. Moreover, some recent studies~\cite{IshibuchiISN18} pointed out that the fairness of some currently prevalent performance metrics depends on their parameter settings. In contrast, high-dimensional data visualisation, which is an important and active research field in modern data analysis, is more effective and intuitive to facilitate the DM to understand the trade-off solutions and thus make a meaningful decision.

In the context of EMO, an ideal visualisation technique is expected to satisfy the following prerequisites~\cite{TusarF15}.
\begin{itemize}
    \item It is able to help the DM identify the Pareto dominance relation between different solutions. This will further facilitate the understanding of the trade-off relationship among objective functions.
    \item It is not only able to preserve the geometric characteristics (e.g. shape and location) of the approximated PF, but also can help the DM understand such information. This will finally lead to an intuitive comparison of the distribution of solutions obtained by different EMO algorithms.
    \item It is able to help the DM identify and exploit the region of interest (e.g. knee regions) thus facilitate the post hoc decision-making.
    \item It is well scalable to the number of dimensions and size of the approximation set without sacrificing its simplicity and computational complexity.
    \item It is resilient to some modifications in the approximation set, e.g. adding or removing some solutions.
\end{itemize}

Comparing to the development of new EMO algorithms, the research on visualisation is relatively lukewarm. Many currently prevalent visualisation methods in EMO are directly derived from the modern data analysis field. Generally speaking, they can be divided into three categories. Interested readers are encouraged to find a recent taxonomy from~\cite{FilipicT18}. The first category aims to show all objective information (e.g. scatter plot, bubble chart~\cite{Ashby00}, parallel coordinate plot~\cite{Inselberg12} and heatmap~\cite{PrykeMN06}). From the cognitive perspective, human beings can only visually understand objects in a two- or three-dimensional space. Given this consideration, the second category mainly aims to implement the visualisation in a low-dimensional space by using some dimensionality reduction methods (e.g. principle component analysis~\cite{Jolliffe11}, multidimensional scaling~\cite{WalkerEF13}, isomap~\cite{TenenbaumSL00}, self-organizing map~\cite{Kohonen90}). Partially similar to the second category, the last one tries to map the original data into a new coordinate system (e.g. radial coordinate visualisation~\cite{IbrahimRMD16}, prosection plot~\cite{TusarF15}, polar coordinate system~\cite{HeY16a}).

In this paper, we mainly aim to overview some currently prevalent visualisation techniques. In particular, we divide them into three different categories according to how data (i.e. population of solutions) is represented. To have a better understanding of the pros and cons of different visualisation techniques, we compare their visualisation results on various data sets. In particular, these data sets are collected by running four state-of-the-art EMO algorithms on the classic DTLZ benchmark test problems~\cite{DebTLZ05}. Note that these test problems have different characteristics and PF shapes.

The rest of this paper is organised as follows. \pref{sec:literature} provides an overview of the currently prevalent visualisation techniques according to our taxonomy. \pref{sec:experiments} compares and analyses different visualisation techniques on various empirical data collected by running four state-of-the-art EMO algorithms on various test problems. \pref{sec:conclusion} concludes this paper and provides some thoughts on future directions.


\section{Overview of State-of-the-art Visualisation Techniques in EMO}
\label{sec:literature}

In this section, we overview some currently prevalent visualisation techniques for visually comparing and understanding Pareto approximation sets in the EMO literature. In particular, these visualisation techniques are classified into three categories according to how data is represented.

\subsection{Visualisation of All Objective Information}
\label{sec:original}

The first category of visualisation techniques aims to reveal all individual objective information of the underlying approximation set. Specifically, scatter plot is the most commonly utilised data visualisation technique that provides a holistic exploration of the population distribution. However, the vanilla scatter plot is only useful in a two- or three-objective case. Although a matrix of scatter plots~\cite{Andrews72}, which is an array of scatter plots displaying all possible pairwise combinations of coordinates, can be used to visualise $m$-dimensional data $(m>3)$, it yields at least $\frac{m(m-1)}{2}$ scatter plots with shared scales. In this case, the information will be too cluttered to provide the DM a holistic exploration of the characteristics of the approximation set, e.g. Pareto dominance relation, PF shape and knee region. Bubble chart is another variant of scatter plot which can visualise at most five-objective approximate set in a three-dimensional Cartesian coordinate system. In particular, the additional two objectives are represented as the sizes of those scatter points and their colours.

Instead of an affine projection of the approximation set into a two- or three-dimensional space as in the scatter plot, parallel coordinate plot (PCP)~\cite{Inselberg12}, the most frequently used visualisation technique in evolutionary many-objective optimisation, projects the data into a coordinate system with $m$ parallel axes. In particular, each poly line in the PCP represents a solution that go across all parallel axes where the intersection at each axis is the corresponding objective value. However, due to the cluttering effect and the interference with crossing lines, it is nontrivial to interpret various characteristics of the approximation set. There have been many efforts devoted to the development of PCP variants~\cite{HeinrichW13}. For example, in order to enable the PCP to interpret the trade-off relation between objectives, Zhen et al.~\cite{ZhenLCP017} proposed a parallel coordinates adjustment method. Specifically, conflicting objectives are gathered together where the correlation between any two objectives are measured by  the Spearman's rank correlation coefficient. Although this method is simple and has shown promising results on some synthetic data, the first axis in the re-ordered PCP is chosen in a random manner. In~\cite{LuHZ16}, Lu et al. proposed to use singular value decomposition (SVD) to calculate the contribution of each objective. Afterwards, the objective associated with the highest eigenvalue is chosen as the first axis in the re-ordered PCP. In~\cite{JohanssonJ09}, Yuan et al. developed an interesting PCP variant that explores the synergy between scatter plot and PCP under the same framework. In particular, by using multidimensional scaling, any two or dimensions in the PCP can be conveniently converted to scatter points and vice versa. Furthermore, there is a brushing tool that allows the user to make selection on both points and line segments conveniently.

Similar to PCP, heatmap~\cite{PrykeMN06} is another frequently used technique for high-dimensional data visualisation. Instead of mapping a solution into parallel axes, heatmap translates the objective value into a colour map where a brighter colour usually represents a higher objective value and vice versa. Unfortunately, heatmap suffers from the same cluttering problem as PCP. In principle, the axes re-ordering techniques used in PCP can be directly transferred to improve the interpretability of heatmap. For example, Walker et al.~\cite{WalkerEF13} proposed to use spectral seriation to reorder objectives in heatmap. In addition, Radar chart~\cite{FilipicT18a} is another equivalent variant of PCP where the coordinate system is changed to a polar coordinate system and $m$ radial axes represent $m$ objectives.

\subsection{Visualisation via Dimension Reduction}
\label{sec:dimension_reduction}

The second category of visualisation techniques aims to transform the high-dimensional data into a lower dimensional space to facilitate the human cognition. In modern data analytics, there are many dimension reduction techniques available to implement such transformation. For example, Walker et al. proposed to use multi-dimensional scaling (MDS) to map the original high-dimensional population into a two-dimensional space where pair-wise distance between any two solutions is expected to be preserved as much as possible. In fact, other prevalent dimension reduction techniques, e.g. principle component analysis~\cite{Jolliffe11}, Isomap~\cite{TenenbaumSL00}, locally linear embedding~\cite{SaulRT03} and Laplacian eigenmaps~\cite{BelkinN01} has been widely used to visualise and analyse high-dimensional data in the machine learning community~\cite{ZhangHW10}. However, their usefulness in the context of multi-objective optimisation is not validated yet. Note that since it is inevitable to lose some information after the dimension reduction, the resulting plots might be distorted to recover the characteristics of the original PF. 

Instead of using dimension reduction techniques from machine learning, Blasco et al.~\cite{BlascoDSM08} proposed a new visualisation technique called level diagrams to visualise the approximation set in a objective-wise manner. More specifically, each diagram is a two-dimensional scatter plot where the horizontal coordinate represents the objective value at the corresponding objective while the vertical coordinate indicates the distance with respect to the ideal point. As claimed by the authors, the level diagrams are able to facilitate the investigation of some of the PF characteristics, i.e. discontinuities, closeness to ideal point and ranges of attainable values. However, since the plots are shown in an objective-wise manner, the level diagrams will end up with many subplots as the matrix of scatter plots.

In~\cite{ChiuB10}, Chiu et al. developed a simple dimension reduction method, called Hyper-Radial Visualisation, to visualise the population in a two-dimensional space by grouping the all objectives into two sets. Although this method is simple, most objective information is directly discarded and the trade-off relationship between objectives is ignored.

\subsection{Visualisation via a Transformed Coordinate System}
\label{sec:transformed}

In principle, this last category share some similarity with the dimension reduction as introduced in~~\pref{sec:dimension_reduction}. They also aim to visualise the original high-dimensional data in a lower-dimensional space. But they try to maintain the original information as much as possible. For example, Ibrahim et al.~\cite{IbrahimRMD16} developed a variant of the classic Radial coordinate visualisation (RadViz)~\cite{HoffmanS02}, called 3D-RadViz, by adding an additional dimension. In particular, this additional dimension represents the perpendicular distance between a solution and a hyper-plane formed by the extreme points of the underlying population. By doing so, the 3D-RadViz is able to provide the information of the convergence of the population. As claimed by the authors, 3D-RadViz can also provide some information about the PF shape and the population distribution.



In~\cite{TusarF15}, Tu\v{s}ar and Filipi\v{c} proposed to use prosection method to visualise a four-dimensional approximation set in a three-dimensional space. Furthermore, they also provide some theoretical analysis about the property of using prosection for preserving most of the characteristics of the original PF. However, it is a bit shame that this technique is not easy to be scalable to a higher-dimensional case.



In~\cite{HeY16a}, He and Yen proposed to use a polar coordinate system (PCS) to represent the approximation set. One of the major advantages of using PCS is the intrinsic dimension is two  so that it is easy to interpret by the DMs. In order to map the population from a Cartesian coordinate system to a PCS, the DMs need to have some prior knowledge of the characteristics of the PF. In~\cite{HeY16a}, they only considered three cases, i.e. linear, convex and concave. Accordingly, they developed an equation to calculate the new coordinates, i.e. radial value and angular value which represent the convergence and distribution respectively, in the PCS. 



\section{Experimental Results and Discussion}
\label{sec:experiments}

In order to have a better understanding of the pros and cons of those visualisation techniques introduced in~\pref{sec:literature}, we choose six representative visualisation techniques (.e. PCP, heatmap, MDS, 3D-Radvis, prosection and polar coordinate system ) and compare their visualisation results on PF approximate sets obtained by four state-of-the-art EMO algorithms (i.e. NSGA-III, MOEA/D, IBEA and HypE~\cite{BaderZ11}). In particular, the classic DTLZ test problems~\cite{DebTLZ05} are chosen to form the benchmark suite. Note that these test problems have various characteristics (e.g. biased distribution and multi-modality) and different PF shapes (e.g. linear, concave, degenerate and disconnected) and have been widely used in the literature~\cite{LiKWCR12,LiFKZ14,LiZKLW14,ChenLY18,LiKD15,LiDZZ17,LiKZD15,LiDY18,LiCMY18,WuLKZ18}. The number of objectives considered in our experiments is set to $m\in\{3,5,8,10,15\}$ while the settings for the population size and the number of generations for different test problems are in~\pref{tab:popsize} and \pref{tab:generations}. Note that these settings follow the standard settings suggested in~\cite{LiDZK15}. Furthermore, to make sure that all plots are in the same scale, we normalise the objective values of the data in each approximation set to the range $[0,1]$. Beyond visual comparisons, we also calculate the inverted generational distance (IGD) and Hypervolume (HV) values of the obtained approximation sets. In particular, we only plot the population with the best IGD value for comparison purpose.

\renewcommand\arraystretch{1.2}
\begin{table}[htbp]
    \centering
    \caption{Settings of Population Size for Different Algorithms}
    \begin{tabular}{c|c|c}
        \hline
        $m$ & NSGA-III, IBEA and HypE & MOEA/D  \\
        \hline
        3   & 92 & 91 \\
        5   & 212 & 210 \\
        8   & 156 & 156 \\
        10  & 276 & 275 \\
        15  & 136 & 135 \\
        \hline
    \end{tabular}
    \label{tab:popsize}
\end{table}

\begin{table}[htbp]
    \centering
    \caption{Number of Generations for Different Test Problems}
    \begin{tabular}{c|c|c|c|c|c}
        \hline
        Test problem & $m = 3$    & $m = 5$   & $m = 8$ & $m=10$ & $m=15$\\
        \hline
        DTLZ1        & 400   & 600   & 750    & 1,000 & 1,500 \\
        DTLZ2, DTLZ5 & 250   & 350   & 500    & 750   & 1,000 \\
        DTLZ3, DTLZ7 & 1,000 & 1,000 & 1,000  & 1,500 & 2,000 \\
        DTLZ4, DTLZ6 & 600   & 1,000 & 1,250  & 2,000 & 3,000 \\
        \hline
    \end{tabular}%
    \label{tab:generations}%
\end{table}%

\subsection{Comparison of Six Visualisation Techniques}
\label{sec:comparisons}

Due to the page limit, we only present some selected results while the complete plots can be found from the supplementary file of this paper\footnote{Supplementary file can be downloaded from \href{https://drive.google.com/open?id=1tFW3IMZlSwa-uw81Gh4h4DVFqZYsLPJ5}{this Google Drive share}.}. In the following paragraphs, we will separately discuss the observations for each of those six visualisation techniques used in this paper.

\subsubsection{Observations on MDS}
\label{sec:MDS}

\begin{figure}[htbp]
    \includegraphics[width=\linewidth]{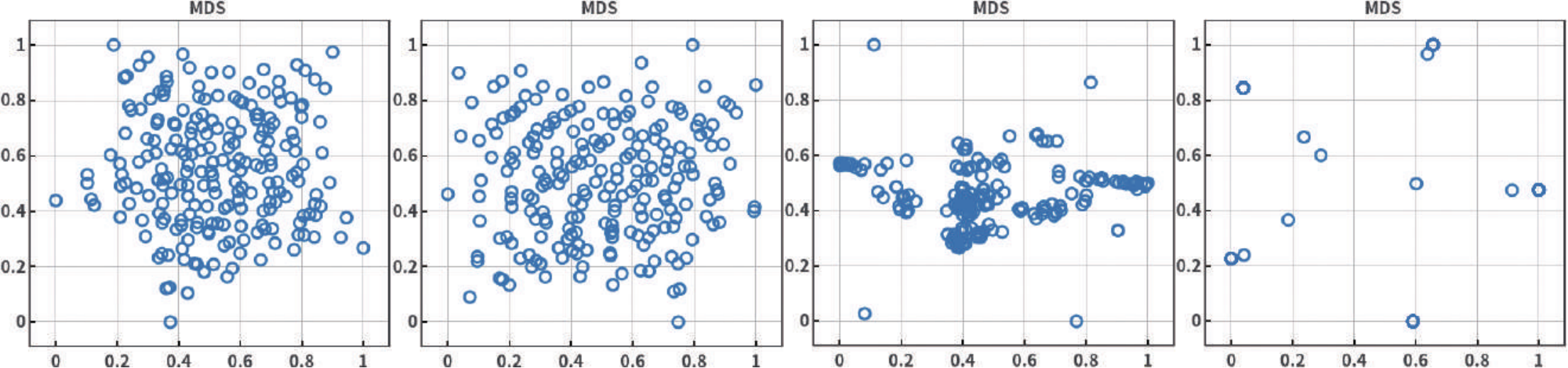}
    \caption{Solutions obtained by NSGA-III, MOEA/D, HypE and IBEA (from left to right) on 5-objective DTLZ1 by using MDS.}
    \label{fig:MDS_DTLZ1M5}
\end{figure}

\begin{figure}[htbp]
    \includegraphics[width=\linewidth]{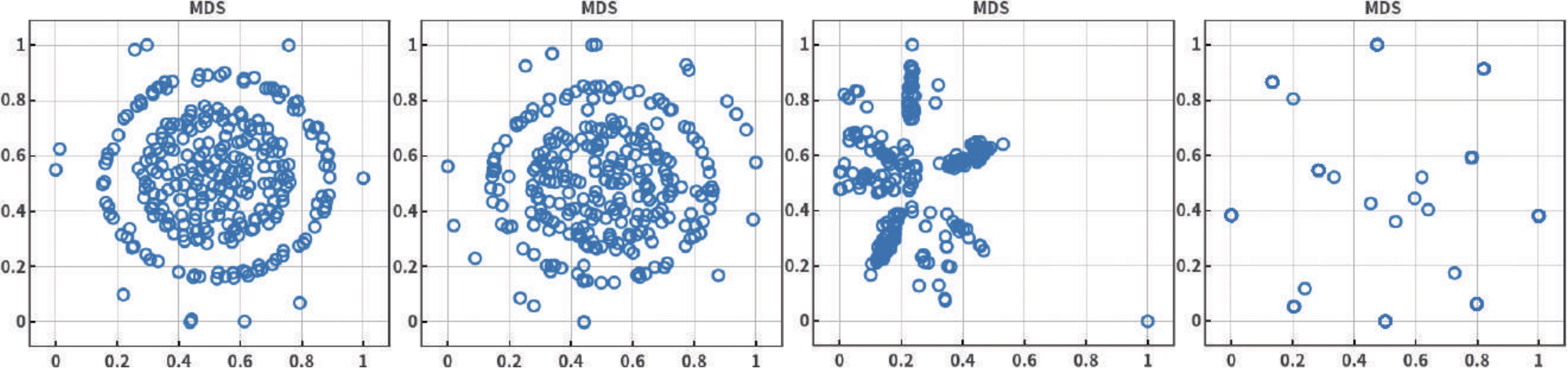}
    \caption{Solutions obtained by NSGA-III, MOEA/D, HypE and IBEA (from left to right) on 10-objective DTLZ1 by using MDS.}
    \label{fig:MDS_DTLZ1M10}
\end{figure}

\begin{figure}[htbp]
    \includegraphics[width=\linewidth]{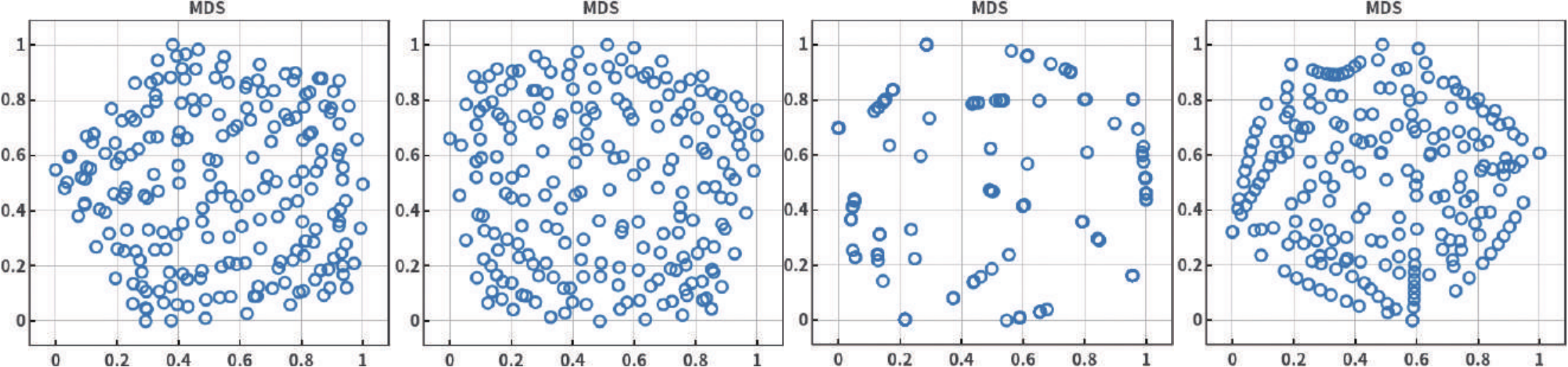}
    \caption{Solutions obtained by NSGA-III, MOEA/D, HypE and IBEA (from left to right) on 5-objective DTLZ4 by using MDS.}
    \label{fig:MDS_DTLZ4M5}
\end{figure}

\begin{figure}[htbp]
    \includegraphics[width=\linewidth]{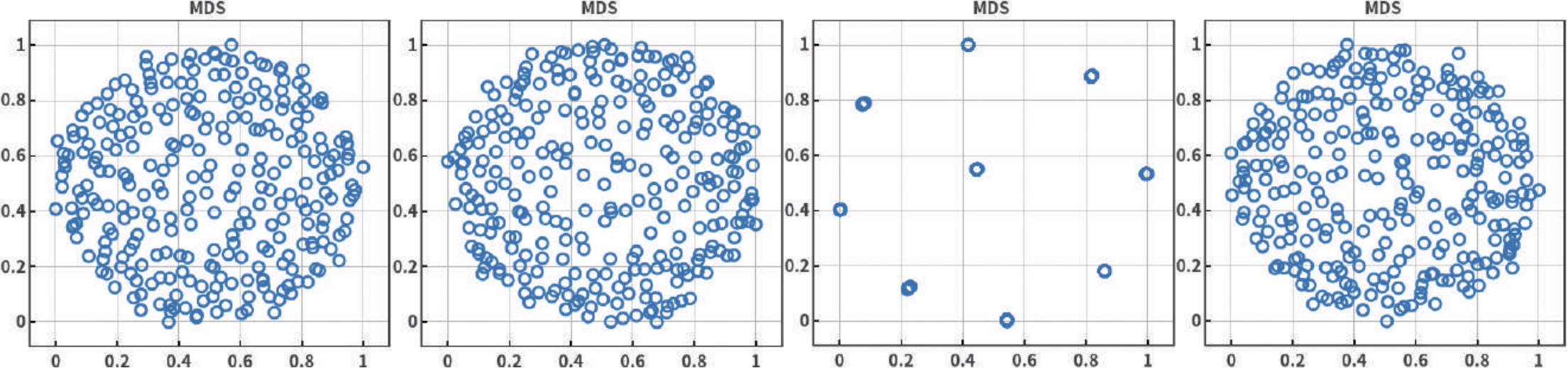}
    \caption{Solutions obtained by NSGA-III, MOEA/D, HypE and IBEA (from left to right) on 10-objective DTLZ4 by using MDS.}
    \label{fig:MDS_DTLZ4M10}
\end{figure}

\begin{figure}[htbp]
    \includegraphics[width=\linewidth]{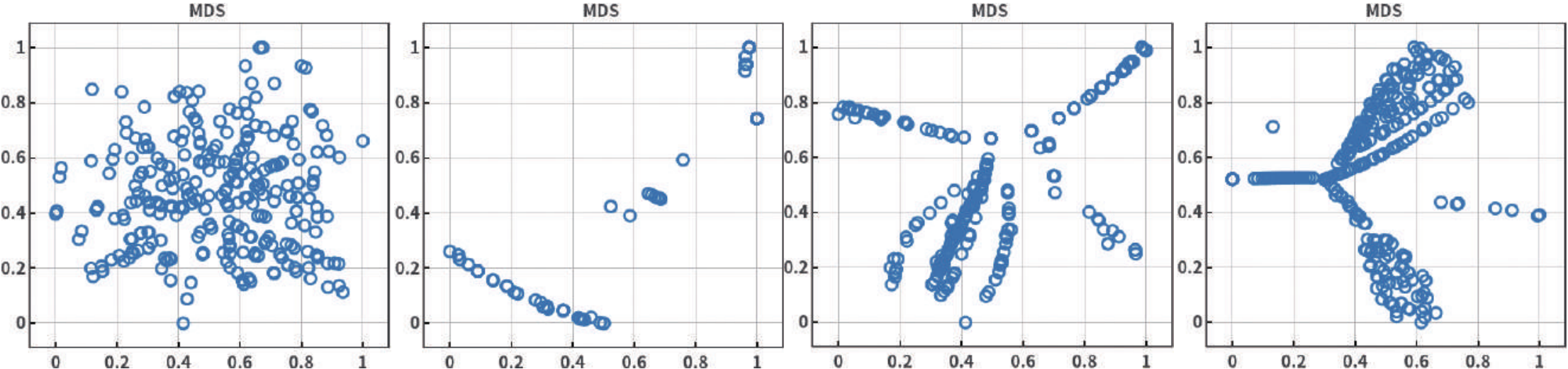}
    \caption{Solutions obtained by NSGA-III, MOEA/D, HypE and IBEA (from left to right) on 10-objective DTLZ5 by using MDS.}
    \label{fig:MDS_DTLZ5M10}
\end{figure}

\begin{figure}[htbp]
    \includegraphics[width=\linewidth]{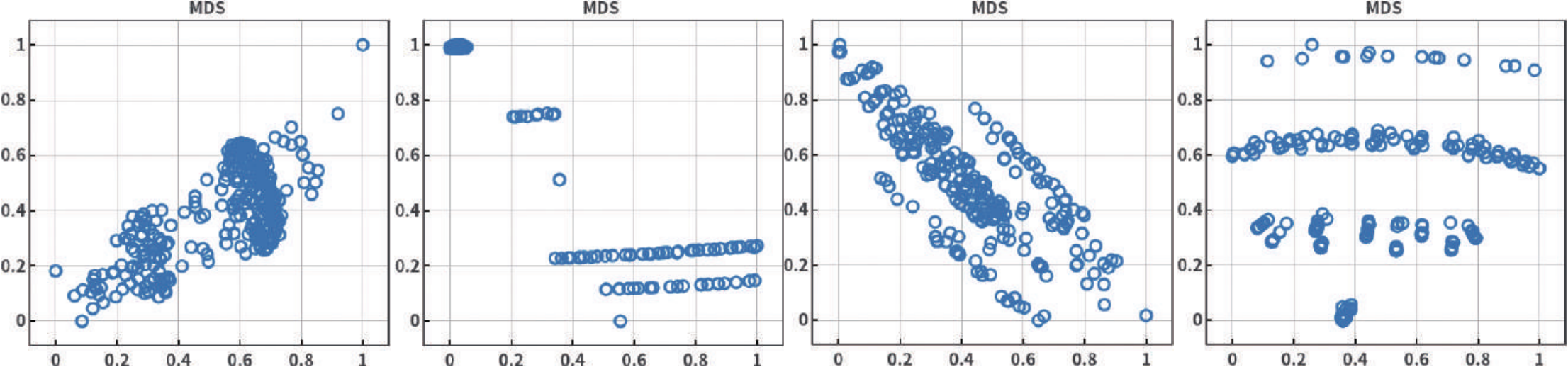}
    \caption{Solutions obtained by NSGA-III, MOEA/D, HypE and IBEA (from left to right) on 10-objective DTLZ7 by using MDS.}
    \label{fig:MDS_DTLZ7M10}
\end{figure}

As shown in~\pref{fig:MDS_DTLZ1M5} to \pref{fig:MDS_DTLZ4M10}, we can clearly see that the performance of NSGA-III and MOEA/D are better than that of HypE and IBEA. Especially for HypE, it can only find some sparsely distributed points along the boundary or extreme of the PF. It is interesting to note that the solution distributions obtained for DTLZ1 and DTLZ4 are visually distinguishable in terms of geometry representations. In particular, the PF shapes of DTLZ1 and DTLZ4 are intrinsically different where the prior one is a convex hyper-plane while the latter one is a concave sphere. However, as shown in~\pref{fig:MDS_DTLZ5M10} and~\pref{fig:MDS_DTLZ7M10}, MDS fails to interpret the geometrical characteristics for some more complex PFs, i.e. DTLZ5 (i.e. degenerate PF) and DTLZ7 (i.e. disconnected PF segments). This might be partially because all algorithms fail to find promising PF approximations on DTLZ5 to DTLZ7. On the other hand, due to the dimensionality reduction, it is inevitable to sacrifice some important information by using the MDS. In this case, the coordinates of those plotted points do not have any physical meaning. Therefore, we can not interpret neither convergence information of the population nor dominance relation by using MDS or other dimension reduction techniques.

\subsubsection{Observations on Prosection}
\label{sec:prosection}

\begin{figure}[htbp]
    \includegraphics[width=\linewidth]{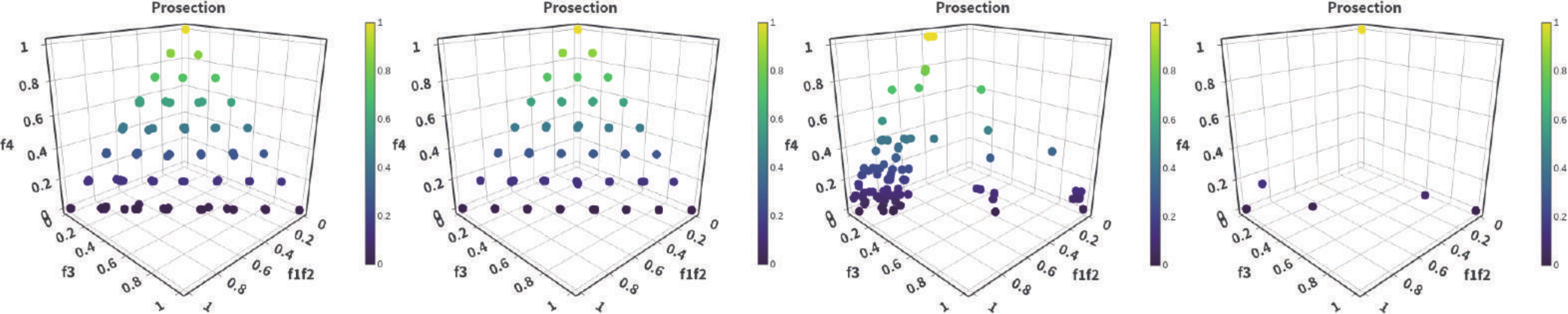}
    \caption{Solutions obtained by NSGA-III, MOEA/D, HypE and IBEA (from left to right) on 4-objective DTLZ1 by using MDS.}
    \label{fig:Prosection_DTLZ1M4}
\end{figure}

\begin{figure}[htbp]
    \includegraphics[width=\linewidth]{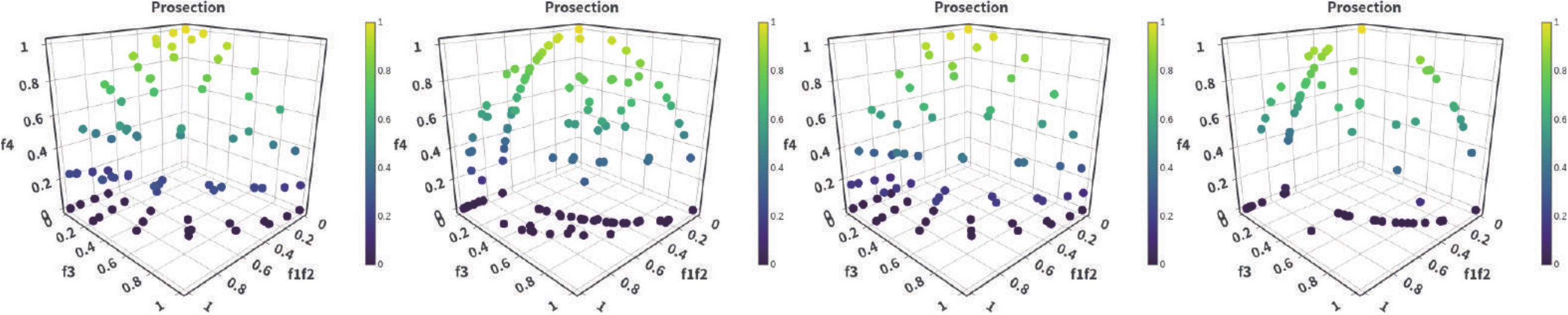}
    \caption{Solutions obtained by NSGA-III, MOEA/D, HypE and IBEA (from left to right) on 4-objective DTLZ4 by using MDS.}
    \label{fig:Prosection_DTLZ4M4}
\end{figure}

\begin{figure}[htbp]
    \includegraphics[width=\linewidth]{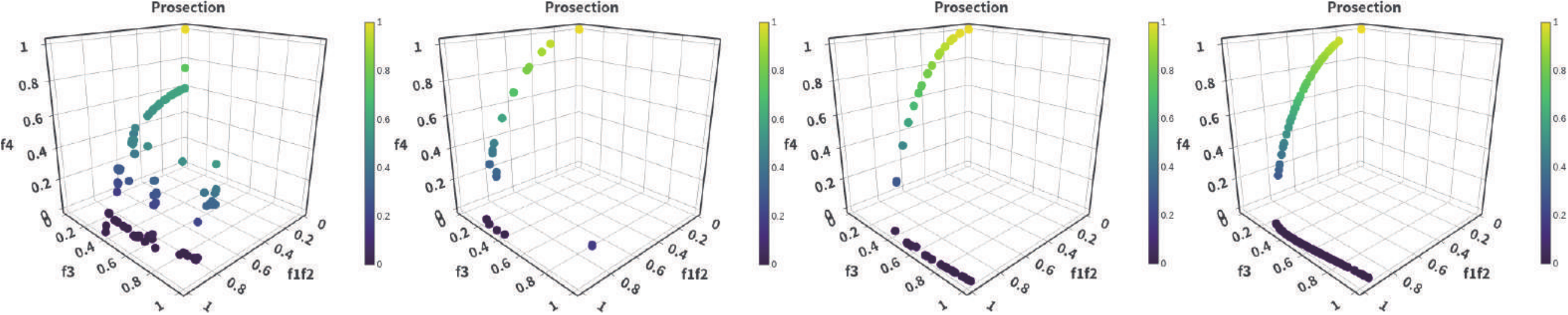}
    \caption{Solutions obtained by NSGA-III, MOEA/D, HypE and IBEA (from left to right) on 4-objective DTLZ5 by using MDS.}
    \label{fig:Prosection_DTLZ5M4}
\end{figure}

\begin{figure}[htbp]
    \includegraphics[width=\linewidth]{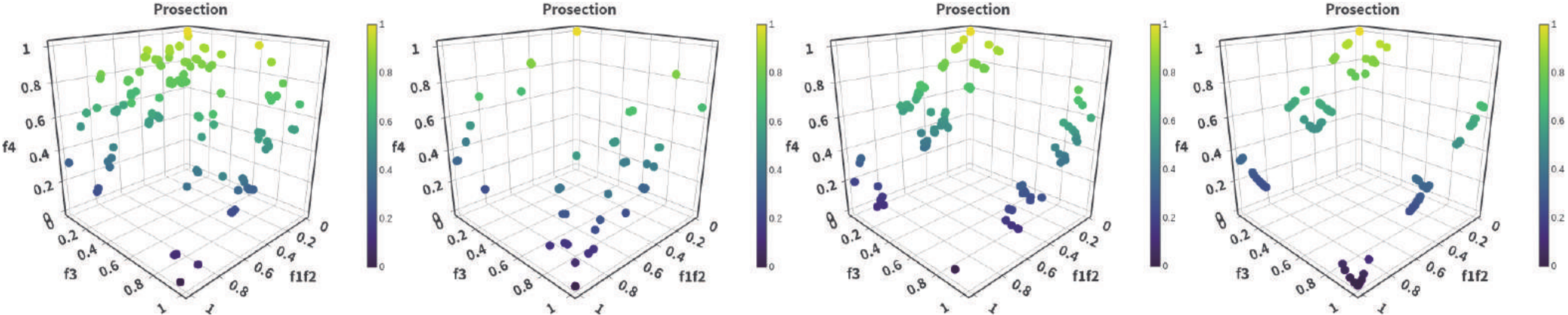}
    \caption{Solutions obtained by NSGA-III, MOEA/D, HypE and IBEA (from left to right) on 4-objective DTLZ7 by using MDS.}
    \label{fig:Prosection_DTLZ7M4}
\end{figure}

In this paper, we set the prosection plain as $f_1$ and $f_2$, the prosection angle as $45\degree$ and the prosection width as 1. As shown in~\pref{fig:Prosection_DTLZ1M4} to~\pref{fig:Prosection_DTLZ7M4}, prosection is able to preserve the geometrical characteristics as we can clearly distinguish the shape with respect to different test problems. Furthermore, since the prosection does not use any dimension reduction technique, the coordinates of different plots have their physical meaning. Therefore, prosection can facilitate the comparison in terms of convergence and dominance relationship. More specifically, as shown in~\pref{fig:Prosection_DTLZ1M4} and~\pref{fig:Prosection_DTLZ4M4}, we can see that NSGA-III and MOEA/D have shown better performance than HypE and IBEA. In particular, as observed in~\pref{fig:MDS_DTLZ1M5} and~\pref{fig:MDS_DTLZ1M10}, IBEA can only find some sparsely distributed points on the boundary or extreme of the PF on DTLZ1. However, as shown in~\pref{fig:Prosection_DTLZ5M4} and~\pref{fig:Prosection_DTLZ7M4}, the performance of NSGA-III and MOEA/D are not as promising as HypE and IBEA. This can be explained as the distribution of weight vectors used in NSGA-III and MOEA/D does not fit the PF shapes of DTLZ5 and DTLZ7~\cite{IshibuchiSMN17,WuLKZZ18}. Note that such observations are not clear when using MDS for visualisation as discussed in~\pref{sec:MDS}. Although prosection has shown very promising results for visually comparing different PF approximation sets on 4-dimensional space, it is not directly scalable to high-dimensional problems.

\subsubsection{Observations on 3D-RadVis}
\label{sec:radviz}

\begin{figure}[htbp]
    \includegraphics[width=\linewidth]{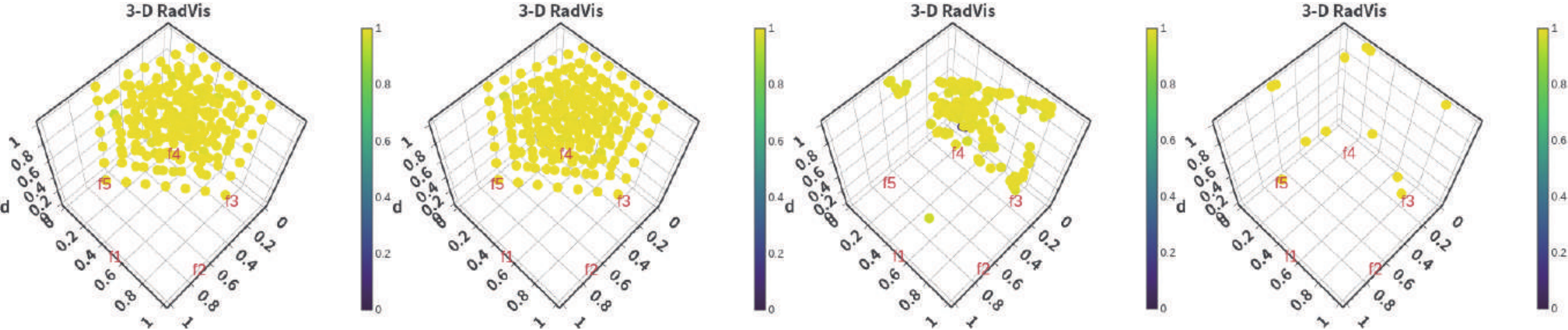}
    \caption{Solutions obtained by NSGA-III, MOEA/D, HypE and IBEA (from left to right) on 5-objective DTLZ1 by using 3D-RadViz.}
    \label{fig:radviz_DTLZ1M5}
\end{figure}

\begin{figure}[htbp]
    \includegraphics[width=\linewidth]{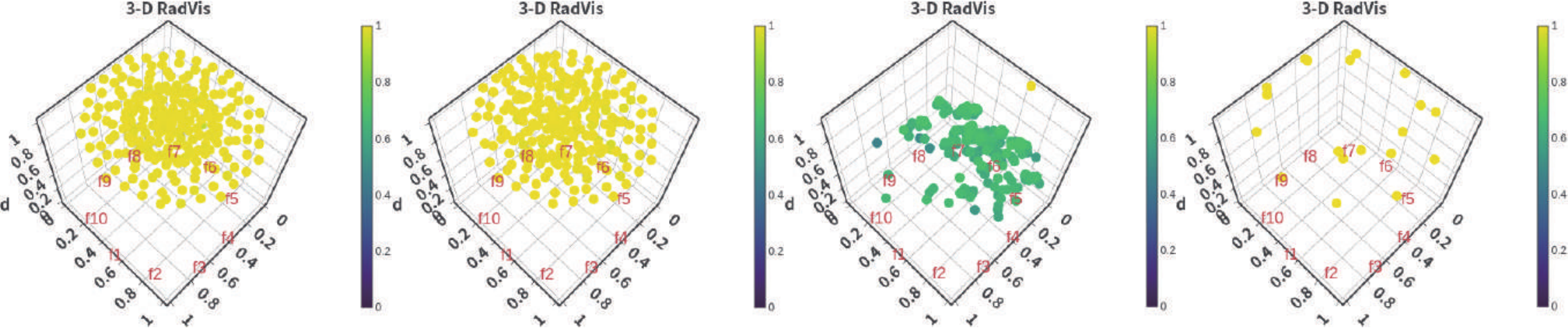}
    \caption{Solutions obtained by NSGA-III, MOEA/D, HypE and IBEA (from left to right) on 10-objective DTLZ1 by using 3D-RadViz.}
    \label{fig:radviz_DTLZ1M10}
\end{figure}

\begin{figure}[htbp]
    \includegraphics[width=\linewidth]{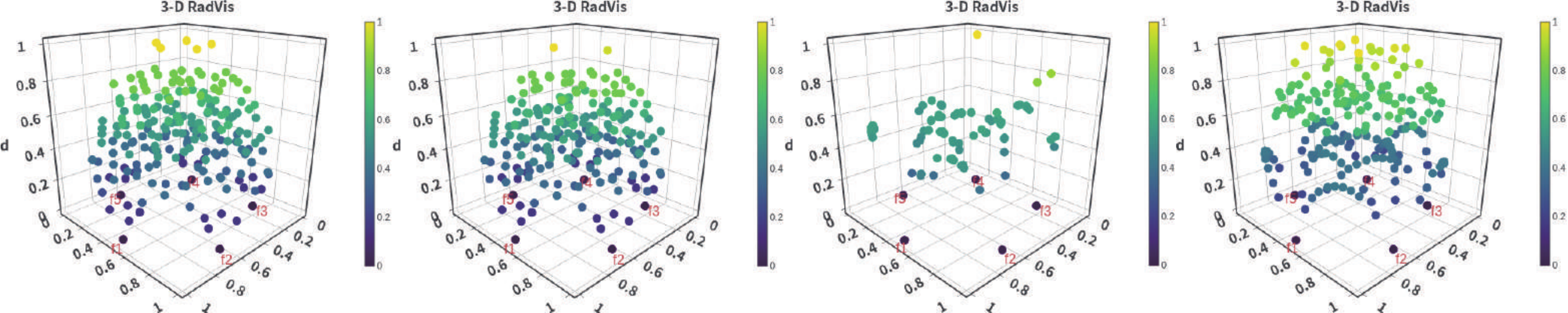}
    \caption{Solutions obtained by NSGA-III, MOEA/D, HypE and IBEA (from left to right) on 5-objective DTLZ4 by using 3D-RadViz.}
    \label{fig:radviz_DTLZ4M5}
\end{figure}

\begin{figure}[htbp]
    \includegraphics[width=\linewidth]{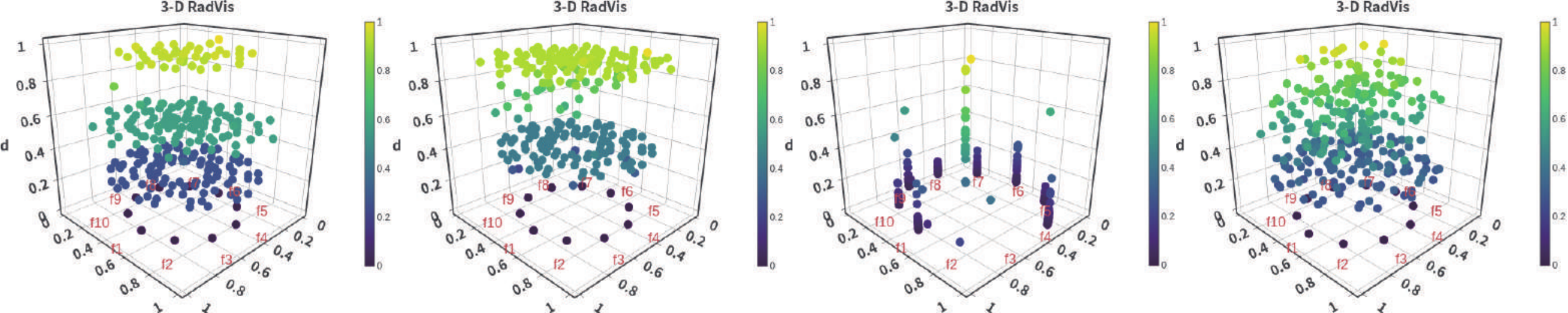}
    \caption{Solutions obtained by NSGA-III, MOEA/D, HypE and IBEA (from left to right) on 10-objective DTLZ4 by using 3D-RadViz.}
    \label{fig:radviz_DTLZ4M10}
\end{figure}

3D-RadViz is a vibrant modification of the original RadViz. Because the additional dimension, which represents the perpendicular distance to a hyper-plane formed by the extreme point of each dimension, 3D-RadViz is able to interpret more information related to the geometry. As shown in~\pref{fig:radviz_DTLZ1M5} to~\pref{fig:radviz_DTLZ4M10}, we can clearly tell the geometrical difference between the solutions with respect to DTLZ1 and DTLZ4. Note that we rotate the axis in order to provide a better view. However, similar to the observations on MDS, the plots on more complex PF shapes (i.e. DTLZ5 and DTLZ7) are not satisfactory where points are cluttered without an explicit regularity. In addition, due to the additional dimension, 3D-RadViz is able to help interpret the convergence of an approximation set but the dominance relationship between solutions is not interpretable.

\begin{figure}[htbp]
    \includegraphics[width=\linewidth]{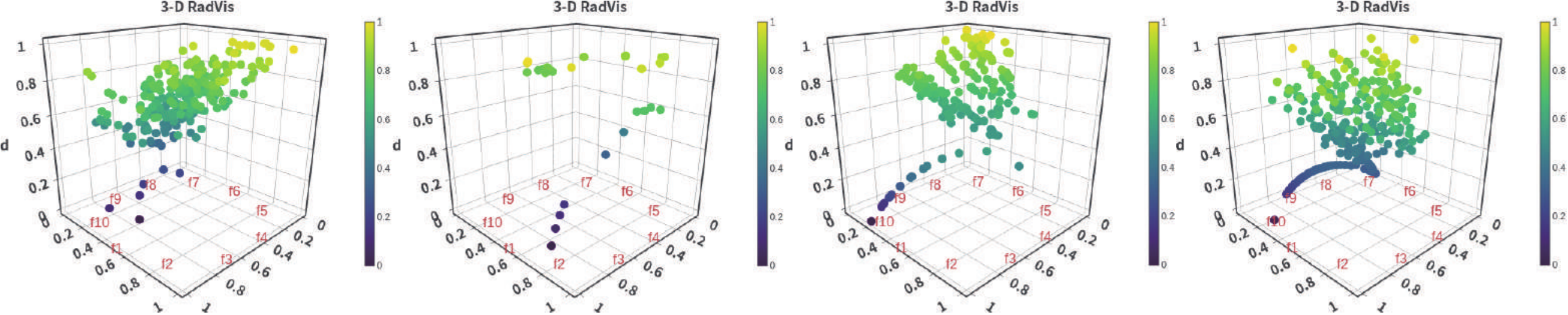}
    \caption{Solutions obtained by NSGA-III, MOEA/D, HypE and IBEA (from left to right) on 10-objective DTLZ5 by using 3D-RadViz.}
    \label{fig:radviz_DTLZ5M10}
\end{figure}

\begin{figure}[htbp]
    \includegraphics[width=\linewidth]{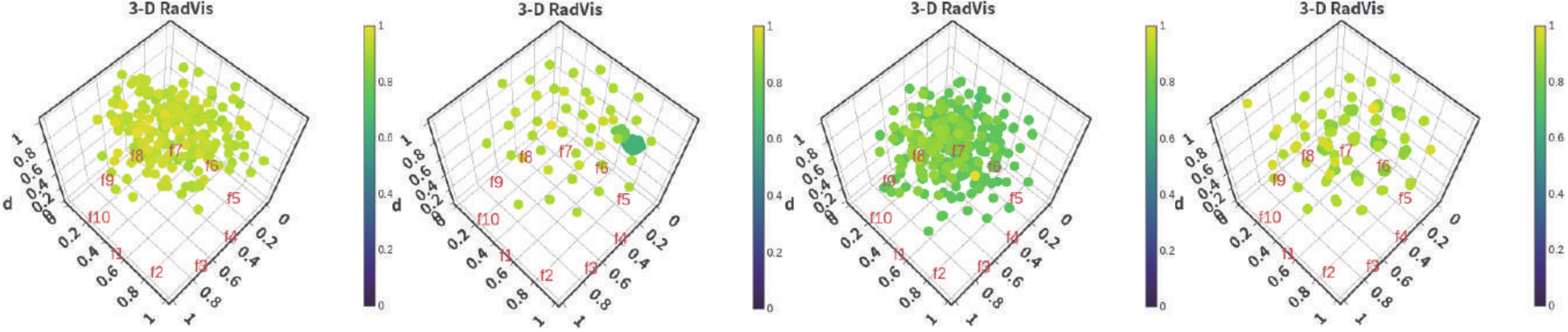}
    \caption{Solutions obtained by NSGA-III, MOEA/D, HypE and IBEA (from left to right) on 10-objective DTLZ7 by using 3D-RadViz.}
    \label{fig:radviz_DTLZ7M10}
\end{figure}

\subsubsection{Observations on PCS}
\label{sec:PCS}

\begin{figure}[htbp]
    \includegraphics[width=\linewidth]{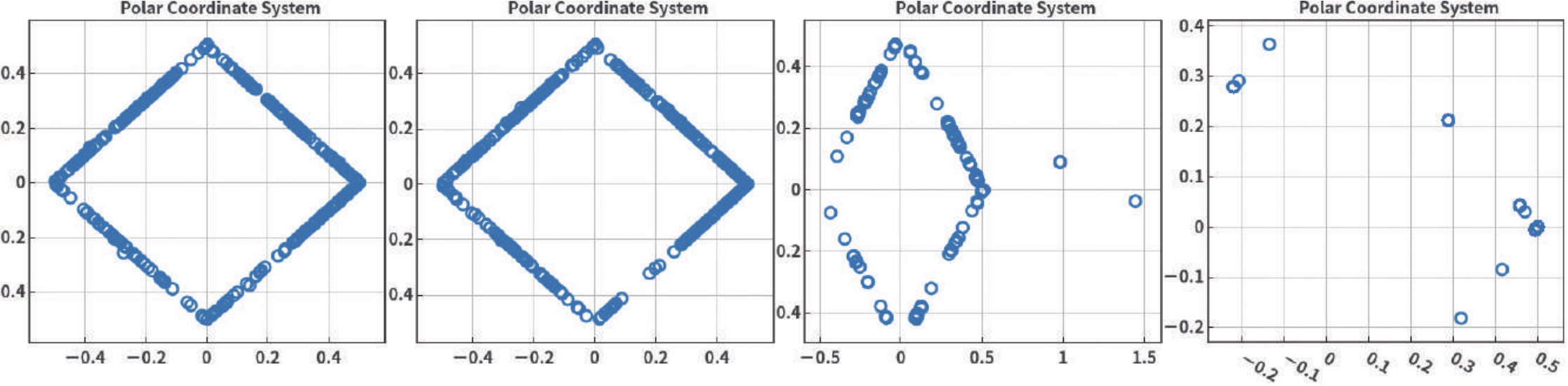}
    \caption{Solutions obtained by NSGA-III, MOEA/D, HypE and IBEA (from left to right) on 5-objective DTLZ1 by using Polar Coordinate System.}
    \label{fig:PCS_DTLZ1M5}
\end{figure}

\begin{figure}[htbp]
    \includegraphics[width=\linewidth]{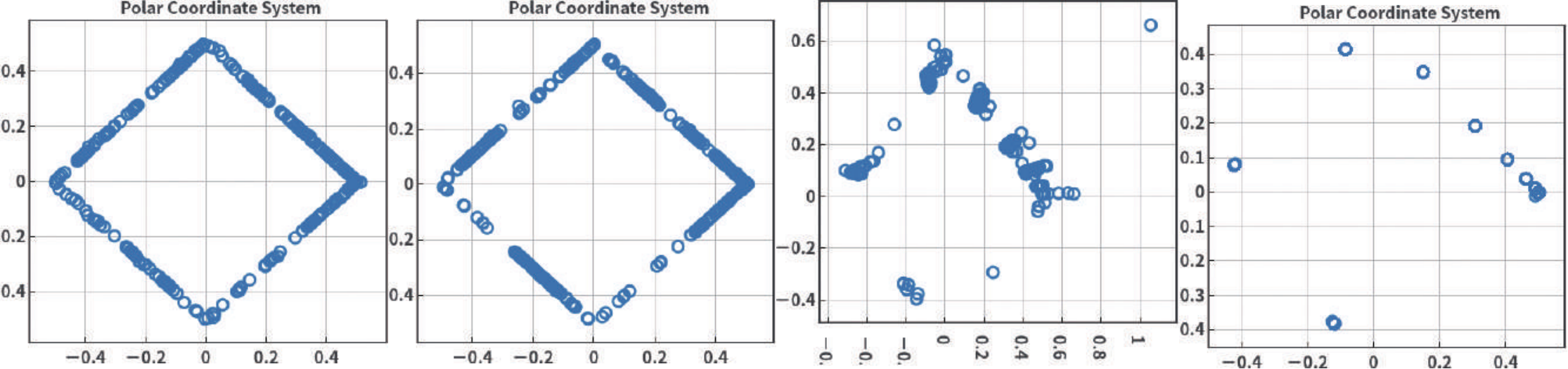}
    \caption{Solutions obtained by NSGA-III, MOEA/D, HypE and IBEA (from left to right) on 10-objective DTLZ1 by using Polar Coordinate System.}
    \label{fig:PCS_DTLZ1M10}
\end{figure}

\begin{figure}[htbp]
    \includegraphics[width=\linewidth]{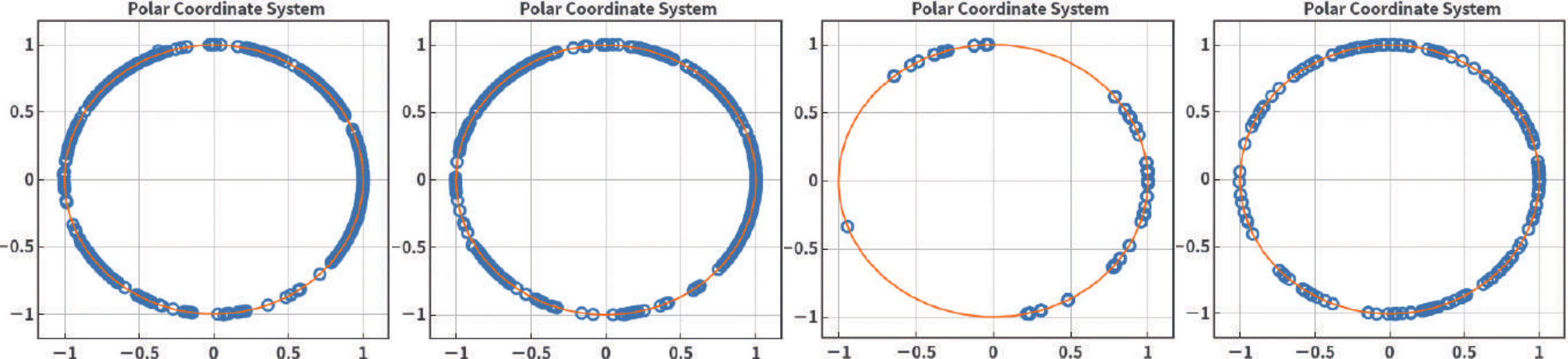}
    \caption{Solutions obtained by NSGA-III, MOEA/D, HypE and IBEA (from left to right) on 5-objective DTLZ4 by using Polar Coordinate System.}
    \label{fig:PCS_DTLZ4M5}
\end{figure}

\begin{figure}[htbp]
    \includegraphics[width=\linewidth]{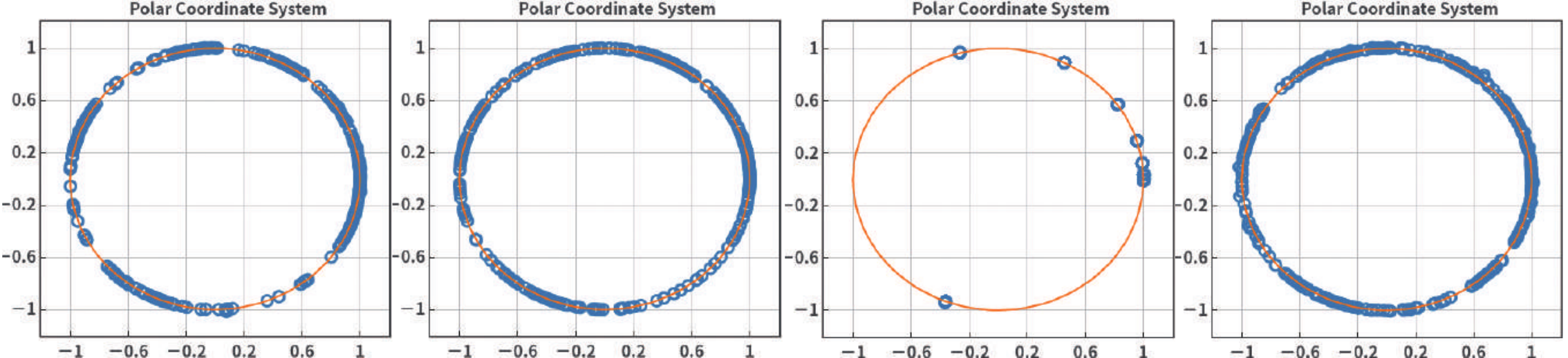}
    \caption{Solutions obtained by NSGA-III, MOEA/D, HypE and IBEA (from left to right) on 10-objective DTLZ4 by using Polar Coordinate System.}
    \label{fig:PCS_DTLZ4M10}
\end{figure}

\begin{figure}[htbp]
    \includegraphics[width=\linewidth]{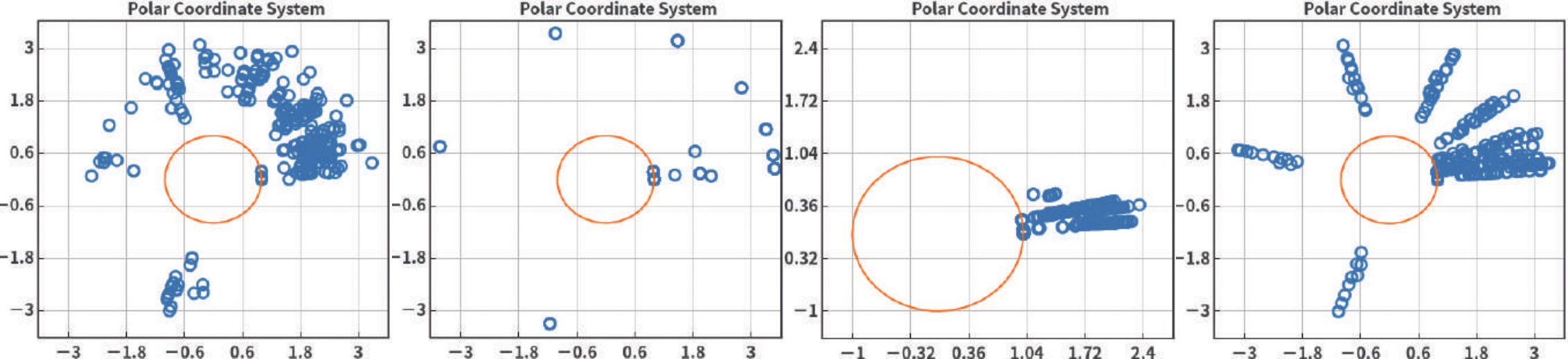}
    \caption{Solutions obtained by NSGA-III, MOEA/D, HypE and IBEA (from left to right) on 10-objective DTLZ5 by using Polar Coordinate System.}
    \label{fig:PCS_DTLZ5M10}
\end{figure}

\begin{figure}[htbp]
    \includegraphics[width=\linewidth]{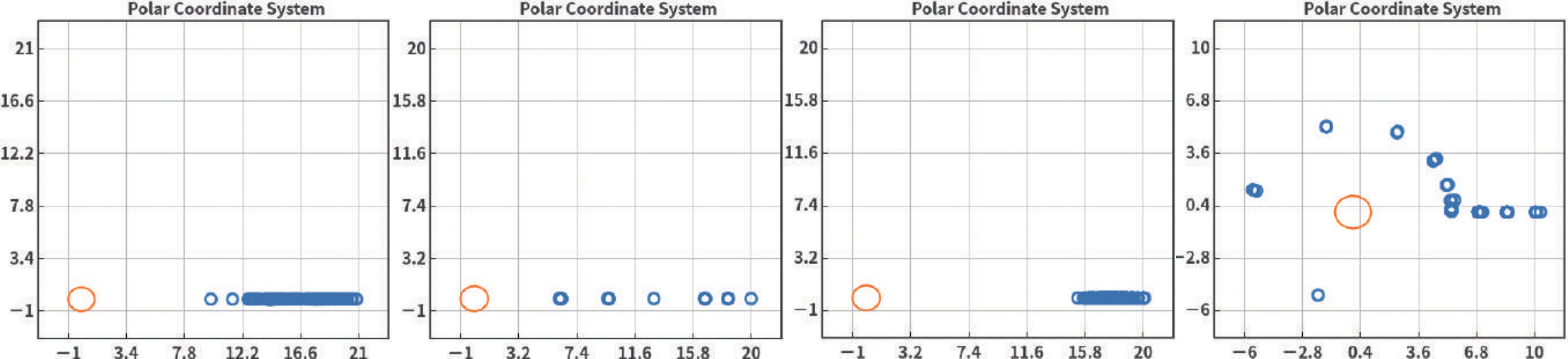}
    \caption{Solutions obtained by NSGA-III, MOEA/D, HypE and IBEA (from left to right) on 10-objective DTLZ7 by using Polar Coordinate System.}
    \label{fig:PCS_DTLZ7M10}
\end{figure}

\pref{fig:PCS_DTLZ1M5} to \pref{fig:PCS_DTLZ7M10} present the results of PCS on various test problems. One of the key characteristics of the PCS is the prior knowledge of the convexity of the target PF. For example, because the PF of DTLZ1 is a convex hyper-plane, the transformed plot in the PCS is a four-edge polygon which is irrelevant to the number of objectives. As shown in~\pref{fig:PCS_DTLZ1M5} and~\pref{fig:PCS_DTLZ1M10}, solutions obtained by NSGA-III and MOEA/D show a good approximation to the target polygon while IBEA only presents some sparsely distributed points. As for DTLZ2, the PF of which is a concave hyper-sphere, the transformed plot in the PCS is an unit circle. As shown in~\pref{fig:PCS_DTLZ4M5} and~\pref{fig:PCS_DTLZ4M10}, NSGA-III, MOEA/D and IBEA obtain a promising approximation to the target circle. In PCS, the closeness to the target plot, either a four-edge polygon or an unit circle, can indicate the convergence of the approximation set while the distribution along the target plot can represent the distribution of the approximation set. However, because the target plot is fixed according to the convexity of the PF itself, it is not easy to interpret the original objective axis in the PCS. Even worse, when the PF does not meet either convex or concave shape, the plotting results are even more difficult to interpret, e.g. \pref{fig:PCS_DTLZ5M10} and \pref{fig:PCS_DTLZ7M10}. Unfortunately, because a real-world optimisation problem is always as a black-box system, it is impossible to have such prior knowledge beforehand.


\subsubsection{Observations on PCP and Heatmap}
\label{sec:PCP}

\begin{figure}[htbp]
    \includegraphics[width=\linewidth]{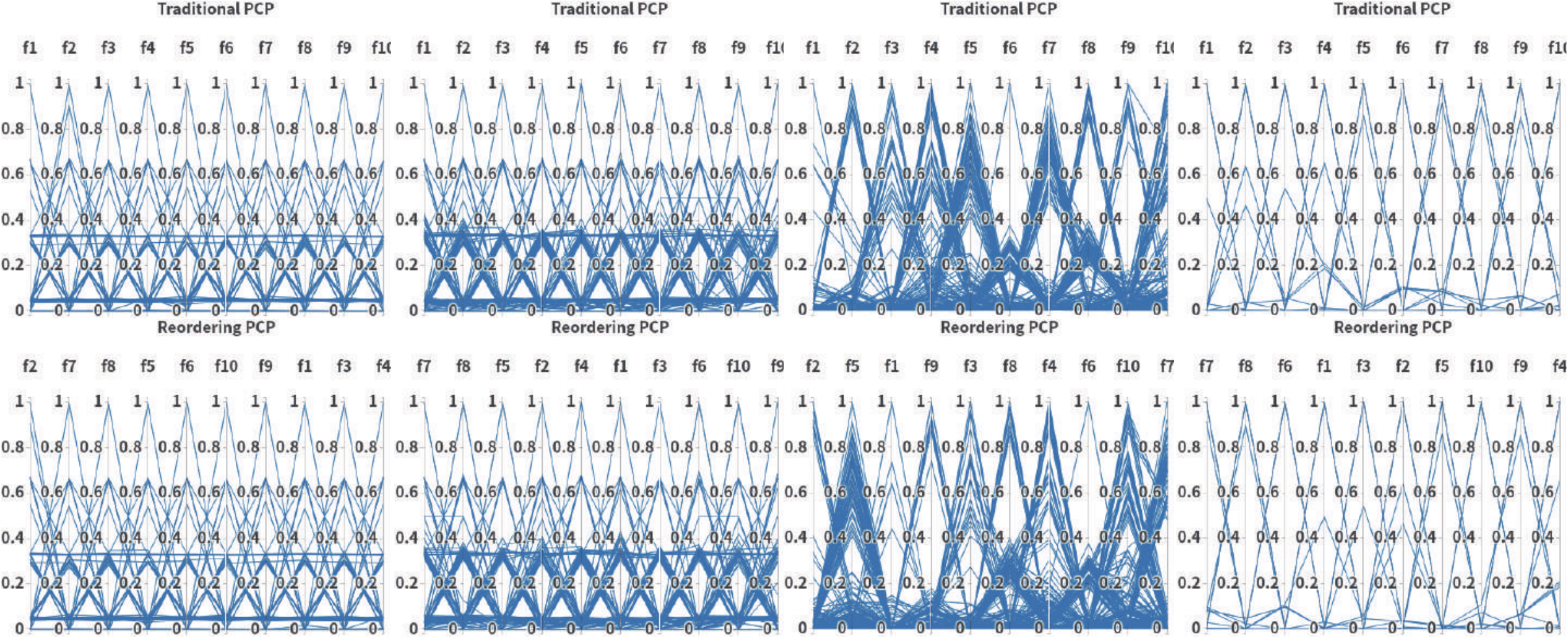}
    \caption{Solutions obtained by NSGA-III, MOEA/D, HypE and IBEA (from left to right) on 10-objective DTLZ1 by using PCP.}
    \label{fig:PCP_DTLZ1M10}
\end{figure}

\begin{figure}[htbp]
    \includegraphics[width=\linewidth]{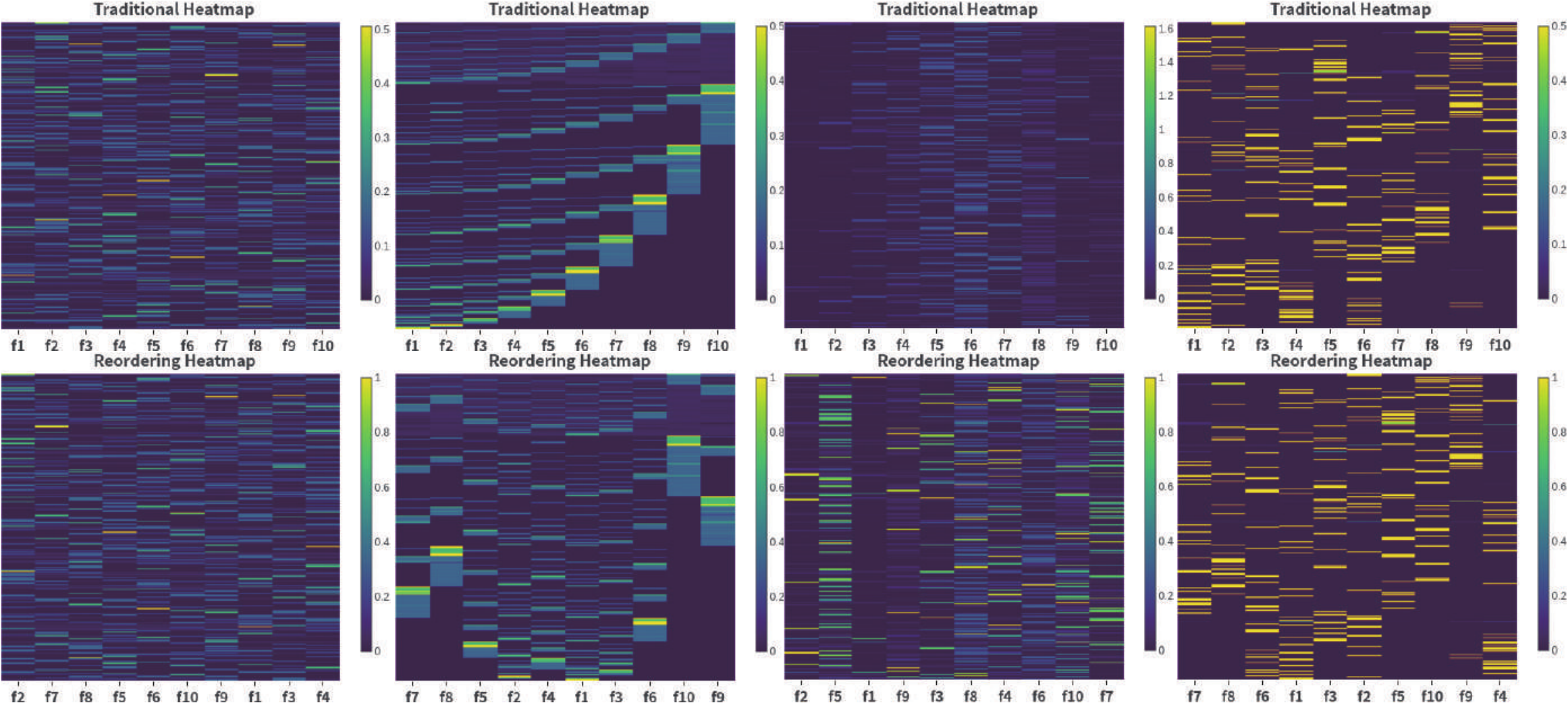}
    \caption{Solutions obtained by NSGA-III, MOEA/D, HypE and IBEA (from left to right) on 10-objective DTLZ1 by using Heatmap.}
    \label{fig:Heatmap_DTLZ1M10}
\end{figure}

\begin{figure}[htbp]
    \includegraphics[width=\linewidth]{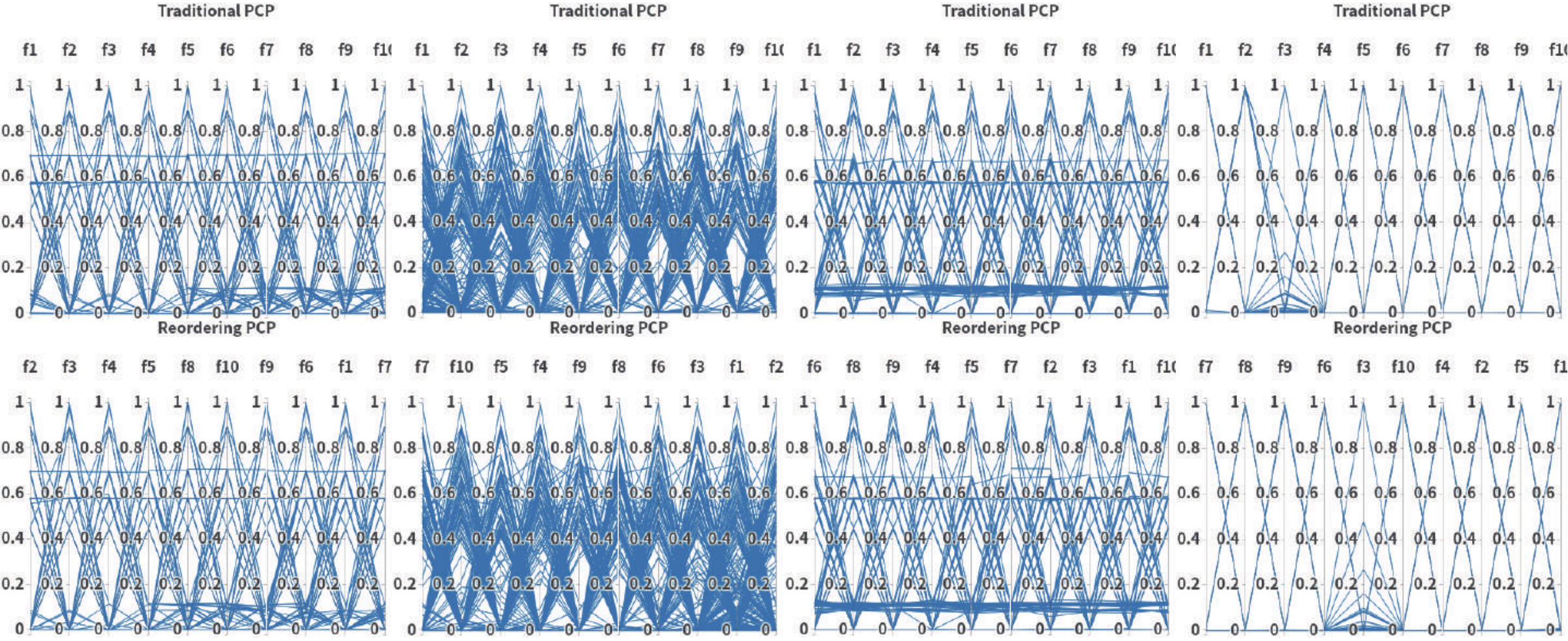}
    \caption{Solutions obtained by NSGA-III, MOEA/D, HypE and IBEA (from left to right) on 10-objective DTLZ4 by using PCP.}
    \label{fig:PCP_DTLZ4M10}
\end{figure}

\begin{figure}[htbp]
    \includegraphics[width=\linewidth]{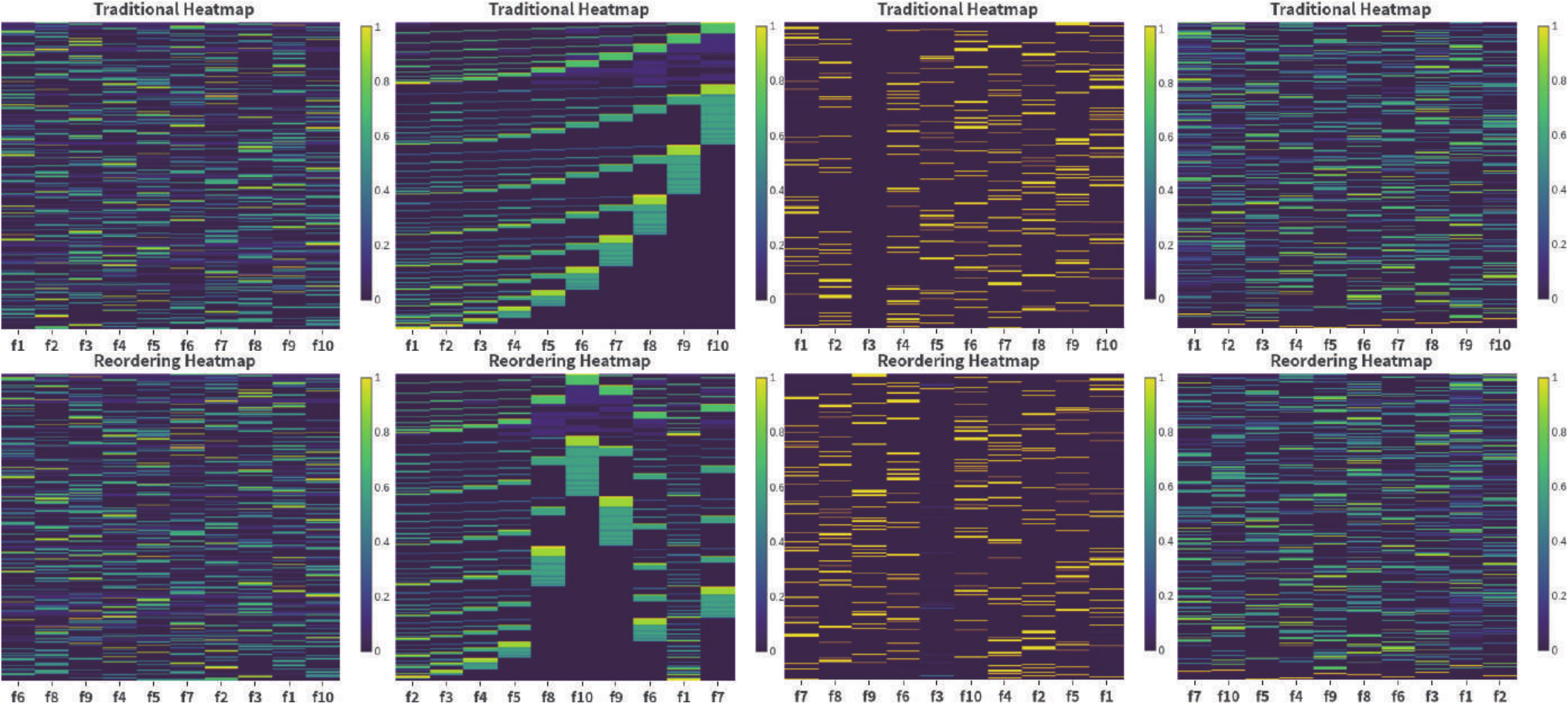}
    \caption{Solutions obtained by NSGA-III, MOEA/D, HypE and IBEA (from left to right) on 10-objective DTLZ4 by using Heatmap.}
    \label{fig:Heatmap_DTLZ4M10}
\end{figure}

\begin{figure}[htbp]
    \includegraphics[width=\linewidth]{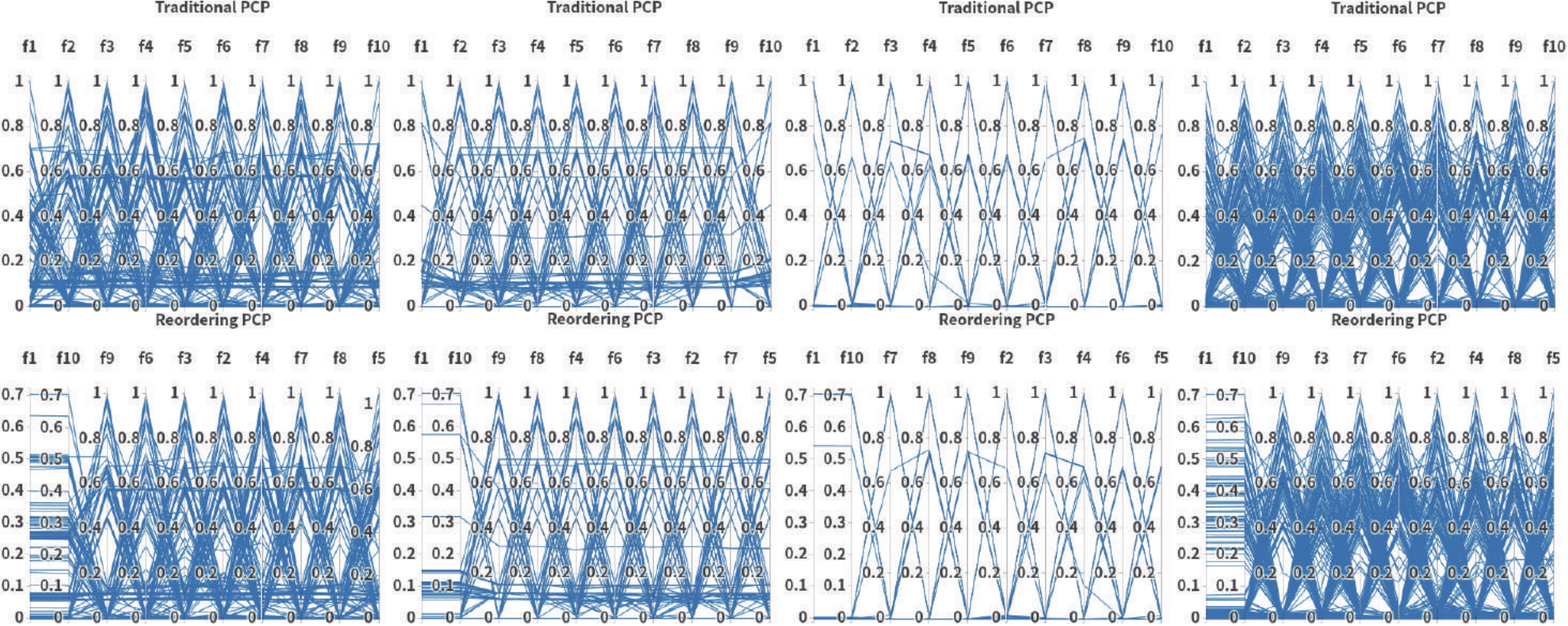}
    \caption{Solutions obtained by NSGA-III, MOEA/D, HypE and IBEA (from left to right) on 10-objective DTLZ5 by using PCP.}
    \label{fig:PCP_DTLZ5M10}
\end{figure}

\begin{figure}[htbp]
    \includegraphics[width=\linewidth]{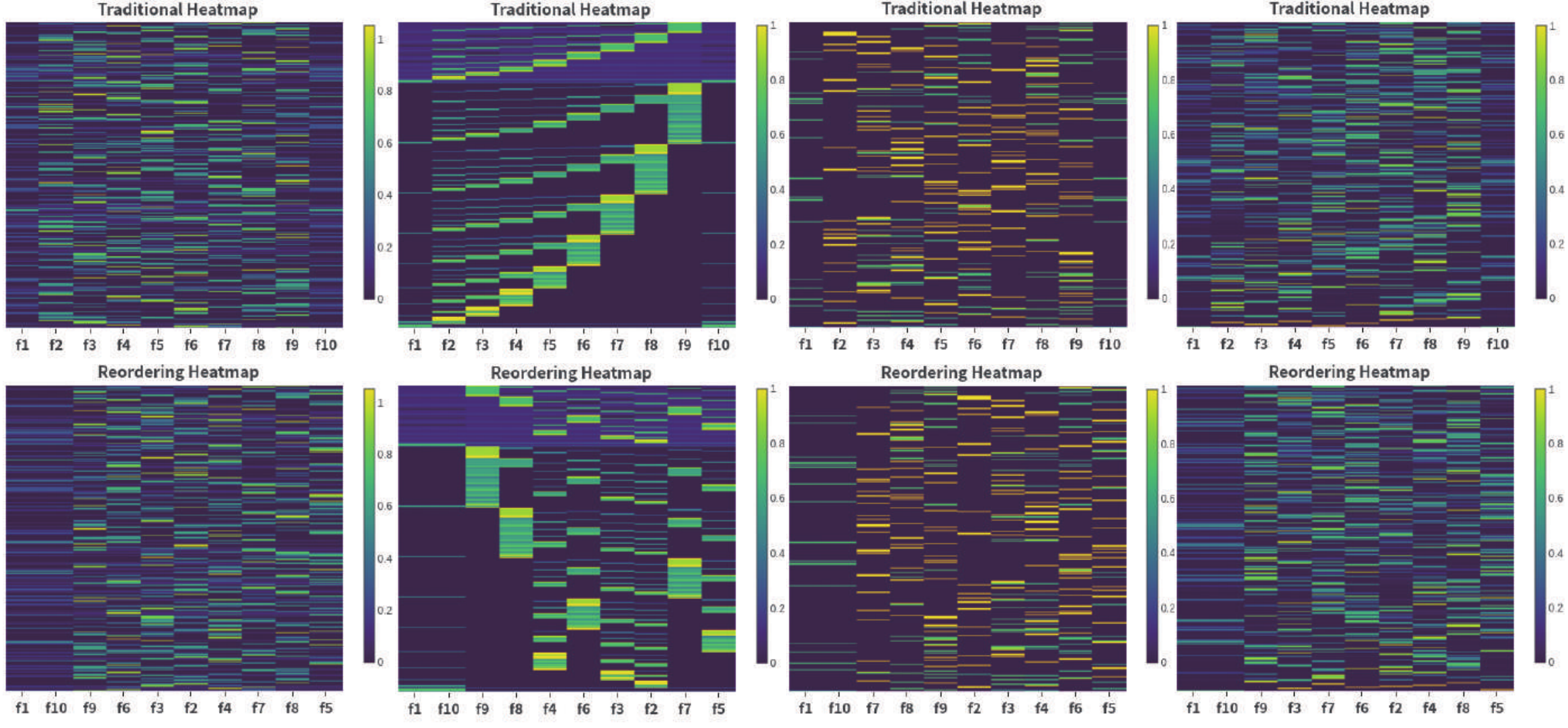}
    \caption{Solutions obtained by NSGA-III, MOEA/D, HypE and IBEA (from left to right) on 10-objective DTLZ5 by using Heatmap.}
    \label{fig:Heatmap_DTLZ5M10}
\end{figure}

\begin{figure}[htbp]
    \includegraphics[width=\linewidth]{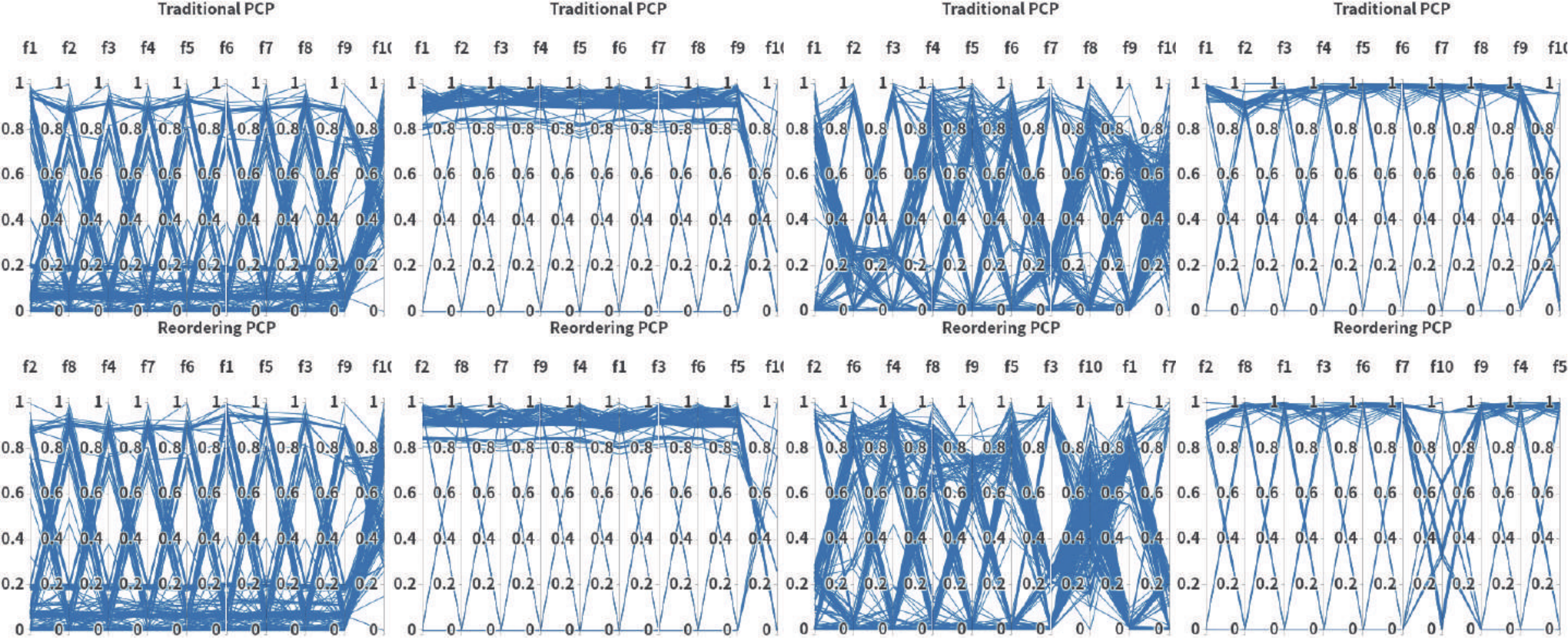}
    \caption{Solutions obtained by NSGA-III, MOEA/D, HypE and IBEA (from left to right) on 10-objective DTLZ7 by using PCP.}
    \label{fig:PCP_DTLZ7M10}
\end{figure}

\begin{figure}[htbp]
    \includegraphics[width=\linewidth]{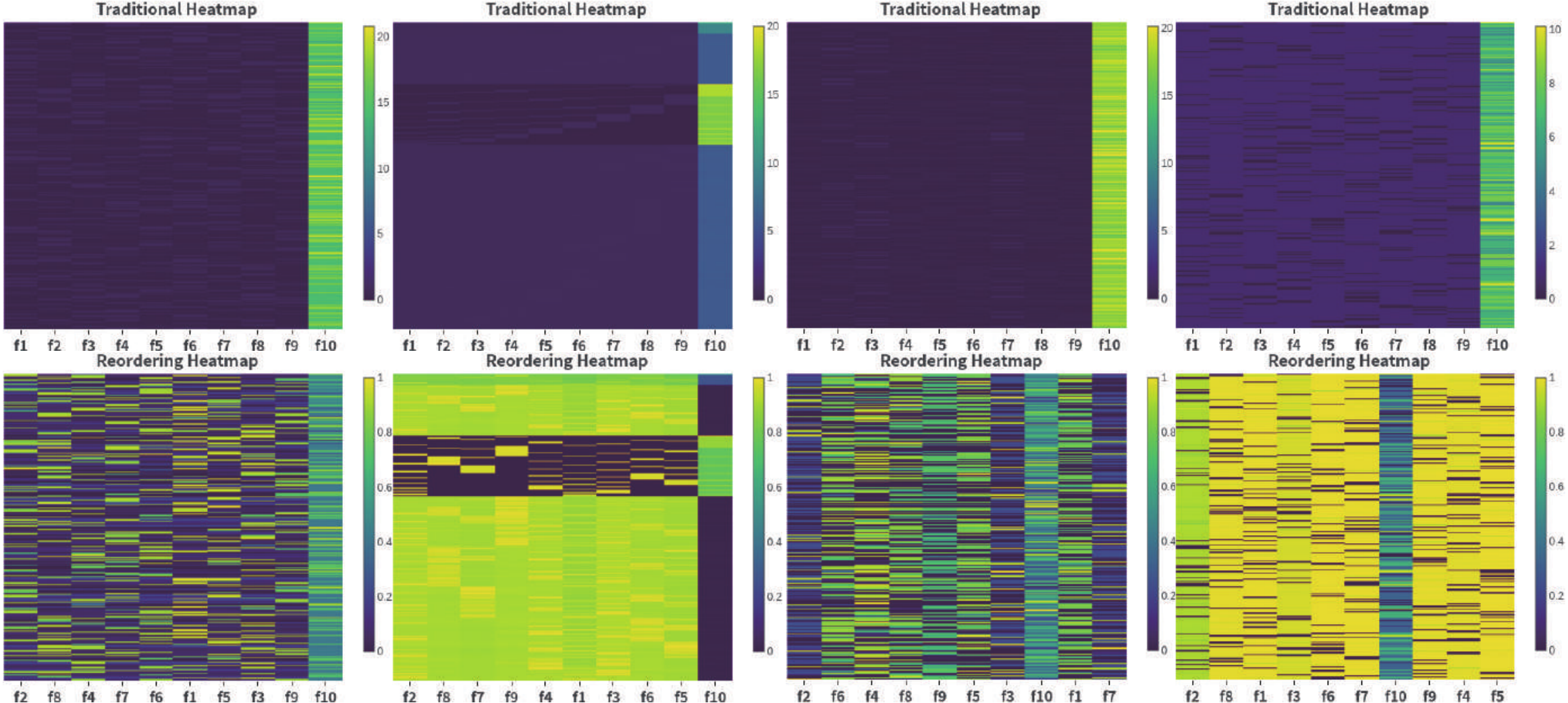}
    \caption{Solutions obtained by NSGA-III, MOEA/D, HypE and IBEA (from left to right) on 10-objective DTLZ7 by using Heatmap.}
    \label{fig:Heatmap_DTLZ7M10}
\end{figure}

As discussed in~\pref{sec:original}, both PCP and heatmap aim to present all objective information in a two-dimensional plot. Because no information is reduced in PCP and heatmap, both of them have the ability to compare the convergence of the approximation set and infer the dominance relationship. But due to the cluttering effect of the original PCP and heatmap, the axis reordering is able to facilitate their interpretability. As shown in~\pref{fig:PCP_DTLZ1M10} to \pref{fig:Heatmap_DTLZ4M10}, we cannot find significant difference between the original PCP and Heatmap with respect to their reordering version. This can be explained as all objectives in DTLZ1 and DTLZ4 are conflicting to each other. In particular, we find that it is easier to investigate the distribution of objective values in heatmap than in PCP. For example, as shown in~\pref{fig:Heatmap_DTLZ1M10} and~\pref{fig:Heatmap_DTLZ4M10}, there are many overlapping solutions obtained by MOEA/D than by NSGA-III. Furthermore, solutions obtained by IBEA on DTLZ1 and HypE on DTLZ4 mainly focus on the extreme region of the PF. In contrast, it is not quite intuitive to read such information from PCP as shown in~\pref{fig:PCP_DTLZ1M10} and \pref{fig:PCP_DTLZ4M10}. DTLZ5 is a degenerate test problem. As shown in~\pref{fig:PCP_DTLZ5M10} and~\pref{fig:Heatmap_DTLZ5M10}, it is much easier to understand that $f_1$ and $f_{10}$ are conflict with each other in DTLZ5 after reordering the axis in PCP and heatmap. Nevertheless, as shown in~\pref{fig:PCP_DTLZ7M10} and~\pref{fig:Heatmap_DTLZ7M10}, both PCP and heatmap are not able to well interpret the disconnected PF segments in DTLZ7.

\section{Conclusions and Future Directions}
\label{sec:conclusion}

This paper overviews some currently prevalent visualisation techniques from the ways of how data is represented. Moreover, in order to have a holistic view of the pros and cons of different types of visualisation techniques, we conduct a series of experiments to visualise several PF approximation sets obtained by four state-of-the-art EMO algorithms on the DTLZ benchmark problems.

From the experimental results, it is interesting to find that the visualisation techniques also follow the \textit{No-Free-Lunch} theorem~\cite{DolpertM97} where no single visualisation technique is able to provide a comprehensive understanding of the characteristics of the approximation set. More specifically, we find that PCP and heatmap are relatively robust for interpreting the convergence and trade-off relationship among objectives. It is interesting to note that the prosection has shown a promising result for understanding both the geometrical characteristics of the approximation set and its convergence. However, it is unfortunately not scalable to more than four objectives. In addition, we cannot interpret the trade-off relationship from the prosection. Dimension reduction techniques like MDS has a potential to interpret the geometrical characteristics of the approximation set. However, due to the information loss, it cannot be used to investigate the convergence and trade-off relationship and it can even be misleading when the PF becomes complex. In principle, the PCS is an interesting idea but due to the requirement of prior knowledge of the convexity of the PF, it is not widely useful when dealing with real-world black-box optimisation problems.

Given these observations, as the next step, we plan to develop an interactive visualisation platform that is able to take advantages of various visualisation techniques under a unified framework. In particular, it is able to progressively interpret various characteristics of the approximation set in a hierarchical manner via an interaction process with DMs. It is also interesting to use the visualisation platform to compare and analysis the performance of more recently proposed EMO algorithms~\cite{LiFK11,LiKWTM13,LiK14}.

\section*{Acknowledgment}
This work was supported by the Royal Society (Grant No. IEC/NSFC/170243).

\bibliographystyle{IEEEtran}
\bibliography{IEEEabrv,vis}

\end{document}